\documentclass[11pt]{article} 
\usepackage[numbers]{natbib}
\usepackage{nyrl,palatino}

\usepackage{graphicx}
\usepackage[utf8]{inputenc} 
\usepackage[T1]{fontenc}    
\usepackage{hyperref}       
\usepackage{url}            
\usepackage{booktabs}       
\usepackage{amsfonts}       
\usepackage{nicefrac}       
\usepackage{microtype}      
\usepackage{xcolor}         
\usepackage{caption} 
\usepackage{algorithm}
\usepackage{algorithmic}
\usepackage{xcolor}         
\usepackage{makecell, multirow} 
\usepackage{colortbl}
\usepackage{booktabs}

\usepackage{subcaption}
\usepackage{graphics,amsmath,amsthm,thmtools,amssymb,enumerate,tikz}

\definecolor{lightergray}{RGB}{230,230,230}

\title{CAAL: Contextual Bandits based Online Hand-Craft Active Learning Strategy Selection}

\author{Shao-An Yin$^{1}$, Jiacong Li$^{2}$, Tianpei Xie$^{2}$, Cecile Levasseur$^{2}$, Wojciech Kowalinski$^{2}$, and Nicola Elia$^{1}$%
\thanks{ This work was conducted while Shao-An Yin was an intern at Amazon.}%
\thanks{$^{1}$Shao-An Yin and Nicola Elia are with the Department of Electrical and Computer Engineering, University of Minnesota, Twin Cities, USA. 
        {\tt\small yin00425@umn.edu, nelia@umn.edu}}%
\thanks{$^{2}$Jiacong Li, Tianpei Xie, Cecile Levasseur, and Wojciech Kowalinski are with Amazon. 
        {\tt\small ljiacon@amazon.com, lukexie@amazon.com, cecilele@amazon.com, kowalin@amazon.com}}}

\begin{document}

\maketitle

\begin{abstract}
The challenge with active learning algorithms is the uncertainty of the statistical distribution of unlabeled data, making it difficult to choose the best hand-crafted strategy. To address this, we introduced \textbf{C}ontextual \textbf{A}daptive \textbf{A}ctive \textbf{L}earning (CAAL). In CAAL, each "arm" represents a hand-crafted strategy. Unlike existing frameworks that select strategies based only on feedback from labeled data, we dynamically choose strategies for labeling batches of data using reward prediction with external context information. This general framework allows for customization with domain knowledge to design more effective rewards and context candidates. In addition, we experimentally show that CAAL outperforms the existing baseline adaptive strategy on public datasets using our reward and context design. Our results are consistent regardless of batch size in each iteration.
\end{abstract}


\startmain 

\section{Introduction}

Active learning aims to minimize labeling effort while maintaining strong model performance~\cite{settles_active_2010}. Most work uses handcrafted strategies to reduce sample complexity~\cite{balcan_true_2010}, with guarantees based on statistical assumptions about the unlabeled data. However, these methods often perform well only for certain data types, which may be unknown in advance.

To address this,~\cite{baram_online_2004} combined multiple handcrafted strategies, and~\cite{Hsu_Lin_2015} extended this idea using an adversarial bandit framework. Yet, adversarial bandits are overly conservative, maintaining constant exploration and often performing similarly to random selection. In active learning, the performance gain from additional labeled data can serve as contextual information. We leverage this by incorporating context into bandit methods, reducing conservativeness and improving adaptability.

Our contributions are as follows:
\begin{enumerate}
    \item We adopt a meta-learner-based active learning approach, more \textbf{industrial-friendly} than recent deep active learning trends, as deep models often underperform on tabular data~\cite{shwartz-ziv_tabular_2022}, which dominate in industry.
    \item We target \textbf{batch selection} and its robustness to batch size, aligning with real industrial needs, rather than the common single-sample selection focus.
    \item We enable adaptability to the best handcrafted strategies for unknown datasets.
    \item We show that incorporating environmental context for \textbf{reward prediction} improves adaptability over adversarial bandit-based state-of-the-art methods.
\end{enumerate}

\subsection{Related Work}
Hand-crafted active learning strategies typically fall into two categories: distribution-based, which assume knowledge of the data distribution, and rejection-based, which maintain a pool of classifiers. The former may lack accurate distribution estimates, while the latter can be hard to scale. To address these limits, recent work explores adaptive strategies that choose the most suitable approach for a given dataset. In this section, we first review individual hand-crafted strategies, then discuss adaptive selection for real-world use.

\paragraph{\textbf{Hand-craft Single Strategies}.}
Recent hand-crafted active learning methods~\cite{NIPS2005_340a3904, hoi_batch_2006, dasgupta_hierarchical_2008, huang_active_2014, sener2018active} offer various heuristics but lack broadly applicable guarantees, with performance varying across datasets. A key reason is their neglect of the sequential feedback nature of active learning, motivating data-driven, adaptive, meta-learning approaches to select better strategies. Rejection-based methods~\cite{NIPS2005_340a3904} were the first to provide theoretical guarantees without strong distributional assumptions, but their need to maintain a pool of base models at each iteration makes them impractical for large-scale training.

\paragraph{\textbf{Deep Active Learning for Batch Selection}.}
Several studies \cite{NEURIPS2019_95323660,Ash2020Deep,ash2021gone,citovsky2021batch} propose handcrafted deep learning-based active learning methods for batch selection, mainly on image data. Yet, deep learning often underperforms tree-based models on tabular datasets \cite{shwartz-ziv_tabular_2022}, which dominate industrial applications. We therefore focus on meta-learner-based batch selection, compatible with any base classifier rather than restricted to deep learning.

\paragraph{\textbf{Data Driven Active Learning}.}
The bandit framework, common in online and adaptive learning, models an agent choosing from a set of actions (arms) with unknown rewards. The agent alternates between \emph{exploration}—sampling arms to estimate rewards—and \emph{exploitation}—selecting the arm with the highest estimated reward. Algorithms for reward estimation vary with assumptions on reward structure.

Early work~\cite{10.1007/978-3-540-74958-5_14,baram_online_2004} applied this idea to active learning, treating each arm as a handcrafted strategy under an adversarial reward setting. Later,~\cite{Hsu_Lin_2015,9093390} designed rewards as unbiased estimators of training loss. \cite{NIPS2017_8ca8da41} argued that limiting arms to handcrafted strategies restricts performance, proposing a new strategy that selects only one sample per iteration—impractical for batch mode. Our approach is closest to~\cite{zhang2023algorithm}, which uses Thompson sampling without contextual information to target class imbalance, assuming i.i.d.\ rewards—an unrealistic condition in sequential querying. Under these constraints,~\cite{Hsu_Lin_2015} remains the state-of-the-art for batch mode active learning.

Building on~\cite{Hsu_Lin_2015}, we propose that the non-stationary \textbf{reward can be predicted from external context}. Unlike the adversarial bandit assumption that treats such rewards as unpredictable, our contextual bandit approach is more greedy and less conservative, yielding better performance.


\section{Contextual Adaptive Active Learning (CAAL)}
In this section, we define the active learning framework and its integration with the bandit setup. Instead of a single predefined strategy, the adaptive framework selects from a pool of human-designed strategies, choosing the most effective based on feedback. We then introduce our \emph{contextual bandit} approach, which predicts each strategy’s reward using external contextual information.

\paragraph{\textbf{Active Learning Framework}.}

\begin{figure}
        \centering
        \begin{subfigure}[b]{0.45\textwidth}
                \includegraphics[width=\textwidth]{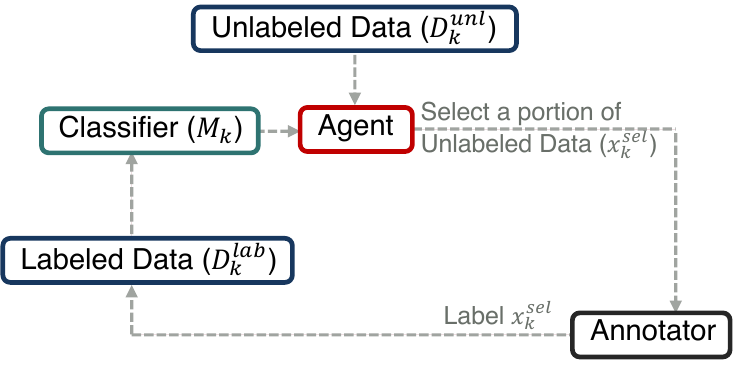}
                \caption{Active Learning Framework.}
        \label{fig:active_learning}
        \end{subfigure}       
        \begin{subfigure}[b]{0.45\textwidth}
                \includegraphics[width=\textwidth]{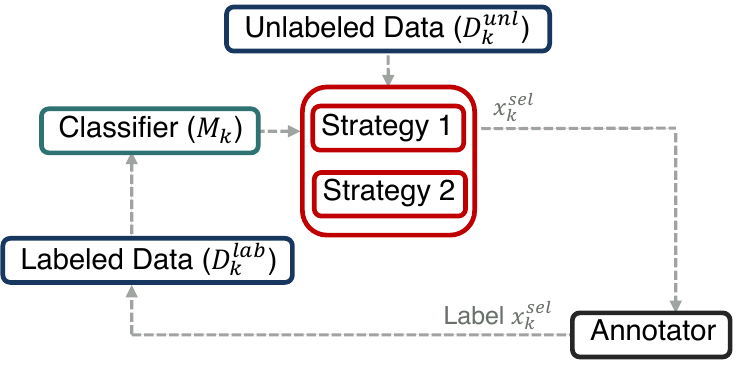}
                \caption{Desired Bandit Based Framework.}
\label{fig:bandit_framework}
\end{subfigure}
\caption{Active learning framework and its adaptive extension for selecting the best strategy.}
\label{fig:lab} 
\end{figure}

Active learning is an iterative process where data is selected from an unlabeled dataset and queried for labels to train a classifier. The goal is to achieve a good classifier with a minimal number of queries. To be precise, at each iteration step $k$, with a given base machine learning classifier $M_k$, labeled dataset $\mathcal{D}^{lab}_k = \{ x^{lab}_k, \, y^{lab}_k\}$, and unlabeled dataset $\mathcal{D}^{unl}_k = \{ x^{unl}_k\}$, a hand-crafted batch mode active learning strategy $\mathcal{S}$ with batch size $b$ is a function that selects $b$ unlabeled data points $x^{sel}_k = \{ x^n \, | \, x^n \in \mathcal{D}^{unl}_k \}_{n=1}^b$ from $\mathcal{D}^{unl}_k$. This selection is based on information gathered from $M_k$ and $\mathcal{D}^{lab}_k$. Therefore, we can express the selection as:
\begin{equation}\label{equ:base_stratey}
    x^{sel}_k = \mathcal{S} \left (M_k, \, \mathcal{D}^{lab}_k, \, \mathcal{D}^{unl}_k \right )
\end{equation}
The annotator will then label the selected dataset $y^{sel}_k$, which will be added to the labeled dataset:
\begin{align}
    &\mathcal{D}^{lab}_{k+1} \leftarrow \mathcal{D}^{lab}_k \cup \{ x^{sel}_k, \, y^{sel}_k\}, \quad
    &\mathcal{D}^{unl}_{k+1} \leftarrow \mathcal{D}^{unl}_k \setminus x^{sel}_k
\end{align}
$\mathcal{D}^{lab}_{k+1}$ is then used to update the base machine learning model $M_{k+1}$, and the entire process repeats. The aim of active learning is to reduce the number of label queries while maintaining a satisfactory performance of the base model. Figure \ref{fig:active_learning} illustrates the framework of active learning.

\paragraph{\textbf{Adaptive framework}.}
Figure \ref{fig:bandit_framework} illustrates the desired adaptive bandit framework. Instead of employing a single hand-crafted active learning strategy defined in equation \ref{equ:base_stratey}, we utilize a collection of such strategies represented by $\mathcal{A}$, analogous to arms in the context of the bandit setup. The adaptive policy $\Pi: \mathbb{R} \to \mathcal{A}$ is a function mapping the context $c_k$ to an action, defined as:
\begin{equation}
a_k = \Pi(c_k).
\end{equation}
Given $r^\star_k$ as the best reward among all arms at time $k$, and $r(a_k)$ as the reward if the agent chooses arm $a$ at time $k$, the experimental regret over $N$ trials of trajectories with a horizon $T$ is defined as:
\begin{equation}
\widehat{\text{Regret}}(\Pi) = \frac{1}{N}\sum_{n = 1}^N \sum_{k=1}^T \left [ r^\star_k - r_n(a_k) \right ] \quad \text{with }a_k = \Pi(c_k).
\end{equation}
The objective of the bandit problem is generally to find a policy function $\Pi$ that minimizes the expected regret.

Assuming the reward is a linear function of the context, i.e., $r(a_k) = \theta_a^\top c_k$, we can leverage the existing LinUCB algorithm \cite{li_contextual-bandit_2010}. However, in the context of active learning, the reward, which we will define explicitly later, is generally non-stationary. Therefore, we can only assume that the reward is locally linear and takes advantage of non-stationary linUCB algorithms, such as discounted linUCB \cite{10.5555/3454287.3455365} (d-linUCB for short) or sliding window linUCB (SW-linUCB for short) \cite{pmlr-v89-cheung19b}.

Our proposed meta-strategy, \emph{Contextual Adaptive Active Learning} (CAAL), operates as follows. At each iteration, the agent receives context from the environment, representing the current state of the active learning process. It then predicts the reward for each arm based on this context and selects the arm with the highest predicted reward. The environment returns the actual reward, and the agent updates its reward prediction function. This process repeats until the end of the trajectory.
\begin{algorithm}
   \caption{Contextual Adaptive Active Learning (CAAL)}
   \label{alg:main}
\begin{algorithmic}[1]
   \STATE {\bfseries Input:} Active Learners ($\mathcal{A}$), Batch Size ($b$), Window Size ($w$). \STATE {\bfseries Initialize:} $\mathbf{C_a} = \mathbf{I}$, $d_a = 0$, $\theta_a = \mathbf{C_a}^{-1} \cdot d_a$ for all arms $a \in \mathcal{A}$.
   \WHILE{True}
   \STATE Observe Context $c_k$ from the environment.
   \STATE Select hand-craft active learner $a_k = \arg \max_{a \in \mathcal{A}} \{ \theta_a^\top c_k + \gamma \sqrt{c_k^\top \mathbf{C_a}^{-1} c_k}\}$.
   \STATE Select $b$ number of unlabeled data $x^{sel}$ based on $a_k$ from $\mathcal{D}^{unl}_k$.
   \STATE Receive $y^{sel}$ from the annotator.
   \STATE $\mathcal{D}^{lab}_{k} \leftarrow \mathcal{D}^{lab}_k \cup \{ x^{sel}_k, \, y^{sel}_k\}$. 
   \STATE $\mathcal{D}^{unl}_{k+1} \leftarrow \mathcal{D}^{unl}_{k} \setminus x^{sel}_k$.
  \STATE Train Classifier $M_k$ with $\mathcal{D}^{lab}_{k}$.
   \STATE Receive Reward $r_k$.
   \STATE $\mathbf{C_{a_k}} \leftarrow \sum_{t = k-w}^k c_t c_t^\top$.
   \STATE $d_{a_k} \leftarrow  \sum_{t = k-w}^k r_t c_t$.
   \STATE $\theta_{a_k} \leftarrow \mathbf{C_{a_k}}^{-1} d_{a_k}$.
   \STATE $k \leftarrow k+1$.
   \ENDWHILE
\end{algorithmic}
\end{algorithm}


\section{Experimental Results}
We design the reward signal and context within this framework, noting that our choice is only one possible design. Experiments are conducted on a real-world dataset for a binary classification task. To ensure fairness, all experiments use logistic regression as the base classifier. We first describe the dataset, then present the evaluation metric aligned with the active learning objective. Finally, we show that our approach outperforms the state-of-the-art adaptive algorithm ALBL on most datasets.

\paragraph{\textbf{Design of Reward and Context}.}

In practice, regardless of how unlabeled data is selected for labeling, an IID control group must be maintained to evaluate model performance. This control group, typically smaller than the training set, provides a reward signal that, to our knowledge, is not commonly used in existing active learning frameworks. We argue that by designing the reward signal from this control group, the reward can depend on external context, enabling our CAAL algorithm to adapt to the best hand-crafted active learning strategy (arm).

Since classifier performance improves with more labeled data, we use dataset size as context. Following \cite{Casanova2020Reinforced}, the reward is the \textbf{relative performance difference} between $M_k$ and $M_{k+1}$, rather than absolute gains, reducing dependence on past arm choices. This design addresses the non-Markovian nature of active learning rewards, mitigates bias, and provides counterfactual information. To handle non-stationarity, we adapt SW-linUCB \cite{pmlr-v89-cheung19b}, modeling rewards as locally linear in context and selecting the arm with the highest predicted improvement under uncertainty.

\paragraph{\textbf{Evaluation Metric}.}
The goal of active learning is to maximize classifier performance while minimizing the number of labeled samples, since labeling is often costly. To evaluate performance, we fix the percentage of queried labeled data and compare the testing ROC-AUC of the resulting model. For fair comparison across datasets, we use the \emph{ROC-AUC ratio} rather than the absolute ROC-AUC, defined as
\begin{equation*}
    \text{ROC-AUC Ratio} = \frac{\text{ROC-AUC}}{\text{ROC-AUC}_{\max}} \times 100 \%,
\end{equation*}
where $\text{ROC-AUC}_{\max}$ is the maximum ROC-AUC obtained using the full training dataset.

To measure the overall performance of a strategy, we further compute the Area Under the Curve (AUC) of the ROC-AUC ratio across query percentages.

\paragraph{\textbf{Dataset}.}
We selected datasets from \cite{zhan_comparative_2021}, which summarizes benchmark datasets in active learning from the UC Irvine Machine Learning Repository. The selection followed two main criteria:
\begin{enumerate}
    \item Sufficiently large to evaluate batch-size active learning strategies.
    \item Sufficiently diverse to highlight performance differences across strategies.
\end{enumerate}

\begin{figure}
  \centering
  \includegraphics[width=1\textwidth]{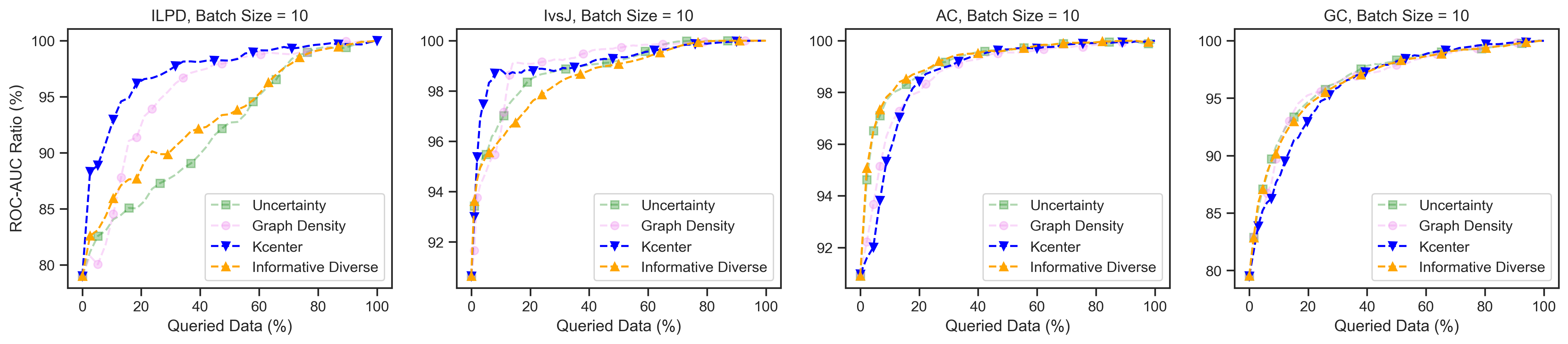}
  \caption{Selected dataset response to active learning strategies with batch size 10.}
  \label{fig:dataset}
\end{figure}

Figure~\ref{fig:dataset} shows that with batch size $10$, handcrafted active learning strategies perform inconsistently across datasets. For example, K-center outperforms others on ILPD and I vs J, while Information Diversity dominates on Australian Credit (AC) and German Credit (GC). Conversely, Information Diversity performs poorly on ILPD and I vs J, whereas K-center lags on AC and GC. We adopt the implementations of these handcrafted strategies from the Google AL toolbox \cite{google_google/active-learning_nodate}.

\begin{table*}
  \centering
\begin{tabular}{l|c|c|c}
\toprule
Dataset & $\#$ of Data & $\#$ of Features & Imbalanced Ratio\\
\midrule
ILPD & 583 & 10 & 2.45\\
IvsJ & 1,502 & 16 & 1.01\\
AC & 690 & 14 & 1.25 \\
GC & 1,000& 20 &2.33 \\
\bottomrule
\end{tabular}
  \caption{Dataset}
  \label{tab:dataset}
\end{table*}

\subsection{Experimental Results}
\begin{figure*}
  \centering
  \includegraphics[width=1\textwidth]{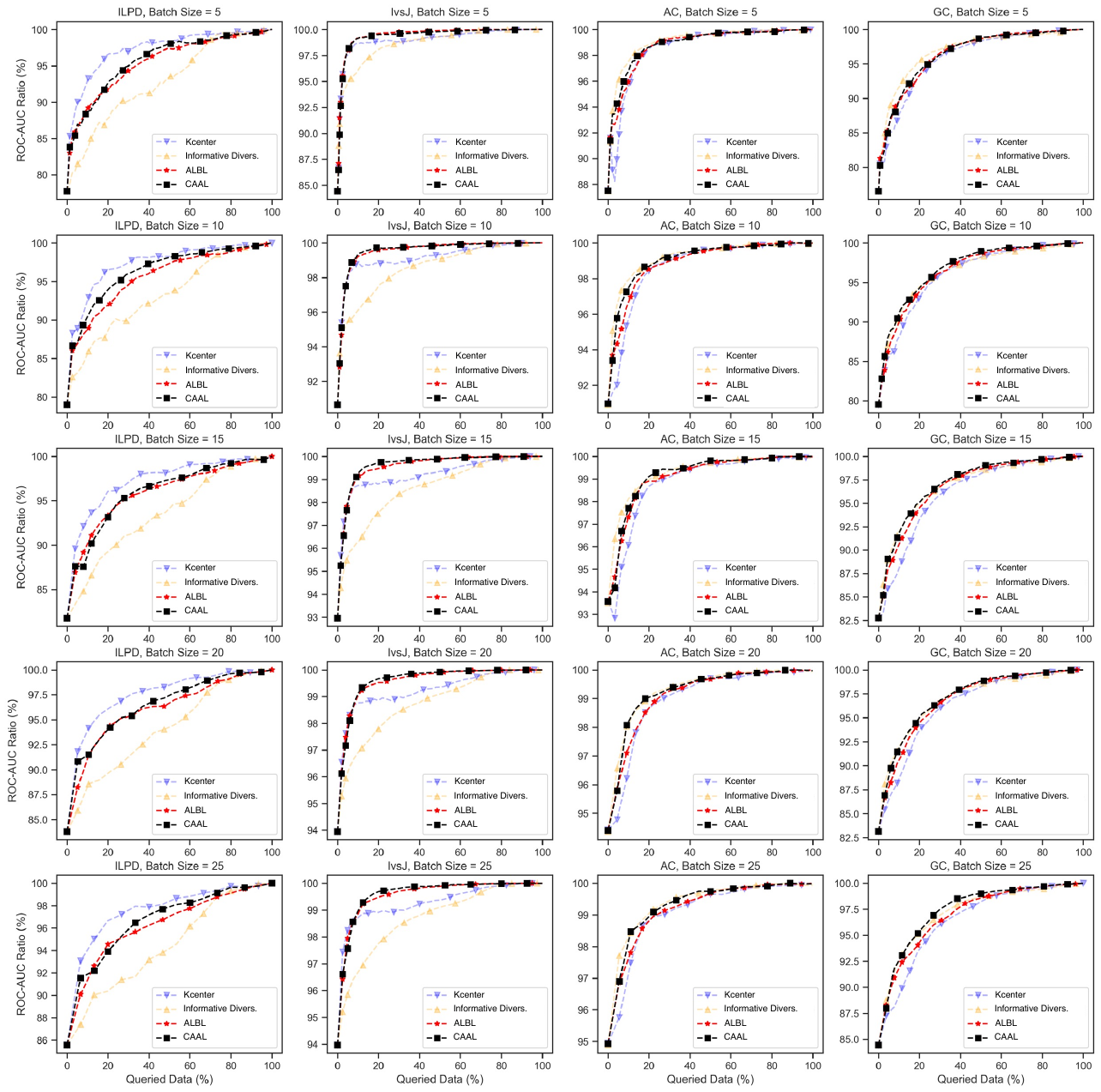}
  \caption{ROC-AUC ratio comparison across batch sizes and datasets.}
  \label{fig:results}
\end{figure*}

Figure~\ref{fig:results} compares K-center, Information Diversity, ALBL \cite{Hsu_Lin_2015}, and our CAAL across batch sizes. The x-axis shows the percentage of queried data, and the y-axis the ROC-AUC ratio on the test set. With batch size $10$, we ran $100$ trials, selecting these hand-crafted strategies for their contrasting behavior. Results indicate that CAAL matches ALBL at batch size $5$ but outperforms it at larger batch sizes, reflecting the less conservative nature of contextual bandits. 

Table~\ref{tab:results} reports the AUC of the ROC-AUC ratio and standard deviations for batch sizes $5$–$25$. Except for size $5$, CAAL consistently achieves better performance, with the small-batch limitation due to linear regression requiring more data. Since real applications often involve larger batches, CAAL is expected to yield stronger gains in practice.

\begin{figure}
  \centering
  \includegraphics[width=1\textwidth]{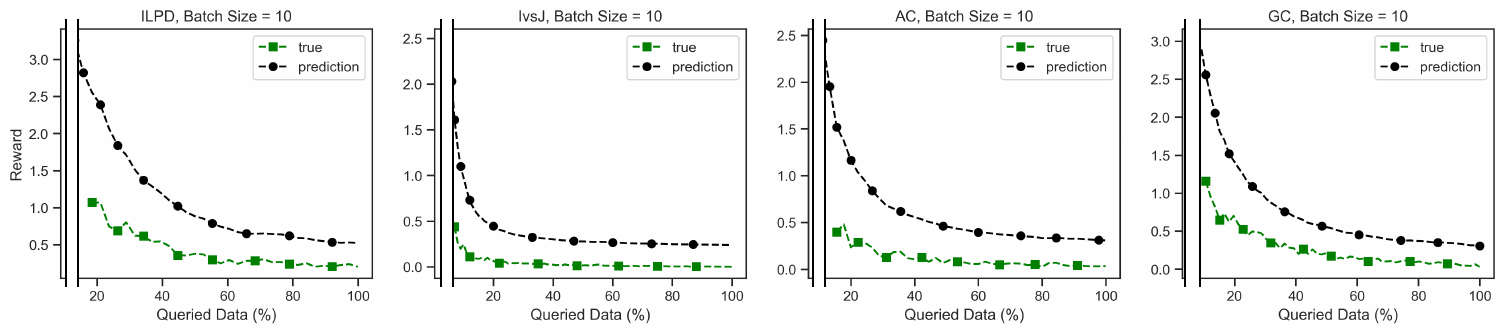}
  \caption{The prediction of reward for batch size of 10.}
  \label{fig:prediction}
\end{figure}

\paragraph{\textbf{Reward Prediction}.}
We study reward prediction from context. Figure~\ref{fig:prediction} shows that linear regression predictions align closely with true rewards, demonstrating that our approach exploits this predictability to avoid the conservative arm choices of ALBL.

\begin{figure}
  \centering
  \includegraphics[width=1\textwidth]{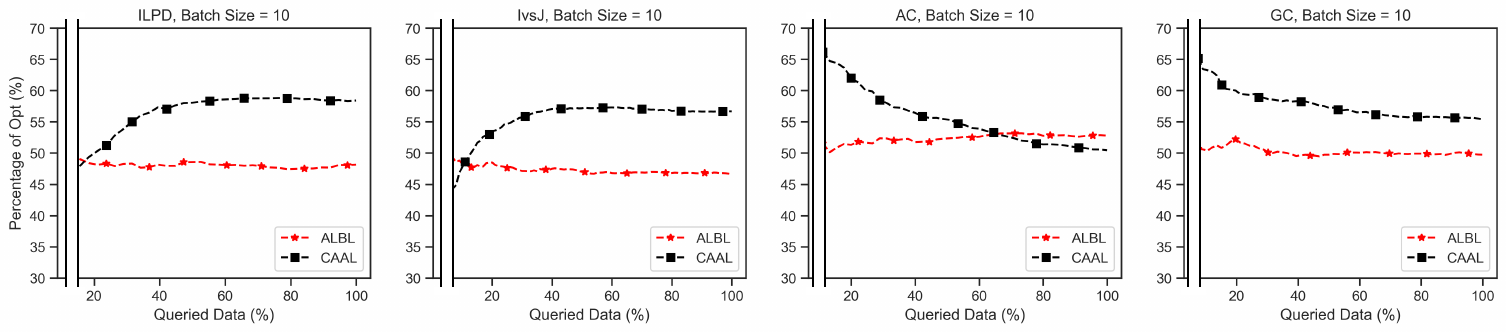}
  \caption{Comparison of the percentage of optimal strategies for batch size of 10.}
  \label{fig:pullhis}
\end{figure}

\paragraph{\textbf{Pull History}.}
Figure~\ref{fig:pullhis} shows ALBL selects arms uniformly ($\sim$50\%), while our method rapidly shifts to the optimal arm and stabilizes after $60\%$ of queries, yielding better performance. 

\begin{table*}
\centering
  \begin{tabular}{l c c|c|c|c|c}
    \toprule
    Data & Strategies &\multicolumn{5}{c}{Batch Size}\\
    \cline{3-7}
    & & \textbf{5}&\textbf{10} & \textbf{15} & \textbf{20} & \textbf{25}\\
    \hline
    ILPD & K-Center&97.20$\pm$0.07&97.11$\pm$0.07  & 97.00$\pm$ 0.07&97.26$\pm$0.07&97.29$\pm$0.07\\
    &Infor.&92.91$\pm$0.08&93.20$\pm$0.08&93.47$\pm$0.08&93.99$\pm$0.08& 94.32$\pm$0.08\\
    &\cellcolor{lightergray}ALBL&\cellcolor{lightergray}95.43$\pm$0.07&\cellcolor{lightergray}95.43$\pm$0.07&\cellcolor{lightergray}95.70$\pm$0.07&\cellcolor{lightergray}95.88$\pm$0.07&\cellcolor{lightergray}96.07$\pm$0.07\\
    &\cellcolor{lightergray}CAAL&\cellcolor{lightergray}\underline{\textbf{95.73}}$\pm$0.07&\cellcolor{lightergray}\underline{\textbf{96.22}}$\pm$0.07&\underline{\textbf{\cellcolor{lightergray}95.80}}$\pm$0.07&\cellcolor{lightergray}\underline{\textbf{96.28}}$\pm$0.07&\cellcolor{lightergray}\underline{\textbf{96.44}}$\pm$0.07\\
    \midrule
    IvsJ& K-Center&99.04$\pm$0.02&99.12$\pm$0.01&99.17$\pm$0.01&99.21$\pm$0.01&99.23$\pm$0.01\\
    &Infor.&98.67$\pm$0.02&98.56$\pm$0.02&98.57$\pm$0.02&98.73$\pm$0.01&98.67$\pm$0.02\\
    &\cellcolor{lightergray}ALBL&\cellcolor{lightergray}\underline{\textbf{99.46}}$\pm$0.02&\cellcolor{lightergray}99.51$\pm$0.01&\cellcolor{lightergray}99.52$\pm$0.01&\cellcolor{lightergray}99.53$\pm$0.01&\cellcolor{lightergray}99.51$\pm$0.01 \\
    &\cellcolor{lightergray} CAAL&\cellcolor{lightergray}99.38$\pm$0.02&\cellcolor{lightergray}\underline{\textbf{99.53}}$\pm$0.01&\cellcolor{lightergray}\underline{\textbf{99.57}}$\pm$0.01&\cellcolor{lightergray}\underline{\textbf{99.56}}$\pm$0.01&\cellcolor{lightergray}\underline{\textbf{99.54}}$\pm$0.01\\
    \midrule
    AC& K-Center&98.55$\pm$0.03&98.65$\pm$0.03&98.81$\pm$0.03&98.92$\pm$0.03&99.01$\pm$0.03\\
    &Infor.&
    99.05$\pm$0.03&99.12$\pm$0.03&99.22$\pm$0.03&99.23$\pm$ 0.03&99.28$\pm$ 0.03\\
    &\cellcolor{lightergray}ALBL&\cellcolor{lightergray}98.76$\pm$0.03&\cellcolor{lightergray}98.83$\pm$0.03&\cellcolor{lightergray}99.04$\pm$0.03&\cellcolor{lightergray}99.06$\pm$0.03&\cellcolor{lightergray}99.13$\pm$0.03\\
    &\cellcolor{lightergray}CAAL&\cellcolor{lightergray}\underline{\textbf{98.84}}$\pm$0.03&\cellcolor{lightergray}\underline{\textbf{99.00}}$\pm$0.03&\cellcolor{lightergray}\underline{\textbf{99.12}}$\pm$0.03&\cellcolor{lightergray}\underline{\textbf{99.18}}$\pm$0.03&\cellcolor{lightergray}\underline{\textbf{99.22}}$\pm$0.03 \\
    \midrule
    GC& K-Center&95.98$\pm$0.05&96.06$\pm$0.05&96.10$\pm$0.05&96.26$\pm$0.05&96.32$\pm$0.05\\
    &Infor.&96.90$\pm$0.05&96.45$\pm$0.05&96.89$\pm$0.05&96.87$\pm$0.04&97.03$\pm$0.04\\
    &\cellcolor{lightergray}ALBL&\cellcolor{lightergray}\underline{\textbf{96.47}}$\pm$0.05&\cellcolor{lightergray}96.46$\pm$0.05&\cellcolor{lightergray}96.76$\pm$0.05&\cellcolor{lightergray}96.74$\pm$0.05&\cellcolor{lightergray}96.82$\pm$0.04 \\
    &\cellcolor{lightergray}CAAL&\cellcolor{lightergray}96.46$\pm$0.04&\cellcolor{lightergray}\underline{\textbf{96.81}}$\pm$0.04&\cellcolor{lightergray}\underline{\textbf{97.06}}$\pm$0.04&\cellcolor{lightergray}\underline{\textbf{96.98}}$\pm$0.04&\cellcolor{lightergray}\underline{\textbf{97.24}}$\pm$0.04\\
    \bottomrule
  \end{tabular}
\caption{Area Under the Curve of the ROC-AUC Ratio. In the table, Infor. represents the Informative Diversity strategy. These are two arms of the adaptive active learning algorithms.}
\label{tab:results}
\end{table*}



\section{Conclusion}
In this paper, we introduced CAAL, an adaptive batch selection algorithm based on contextual bandits. CAAL outperforms the ALBL method on real-world datasets, \textbf{regardless of batch size}, by \textbf{predicting rewards from context}. Future work will explore alternative designs for context and reward signals.



\bibliography{reference}

@String{Computing = "Computing" }

@String{Computer = "{IEEE} Computer" }

@String{Springer = "Springer-Verlag" }

@ArtifactSoftware{R,
    title = {R: A Language and Environment for Statistical Computing},
    author = {{R Core Team}},
    organization = {R Foundation for Statistical Computing},
    address = {Vienna, Austria},
    year = {2019},
    url = {https://www.R-project.org/},
}

@book{settles_active_2010,
	series = {Computer {Sciences} {Technical} {Report}},
	title = {Active {Learning} {Literature} {Survey}},
	number = {1648},
	publisher = {University of Wisconsin–Madison},
	author = {Settles, Burr},
	month = jan,
	year = {2010},
}

@article{balcan_true_2010,
	title = {The true sample complexity of active learning},
	volume = {80},
	issn = {0885-6125, 1573-0565},
	url = {http://link.springer.com/10.1007/s10994-010-5174-y},
	doi = {10.1007/s10994-010-5174-y},
	language = {en},
	number = {2-3},
	urldate = {2024-03-29},
	journal = {Machine Learning},
	author = {Balcan, Maria-Florina and Hanneke, Steve and Vaughan, Jennifer Wortman},
	month = sep,
	year = {2010},
	pages = {111--139},
}

@inproceedings{NIPS2005_340a3904,
 author = {Gilad-bachrach, Ran and Navot, Amir and Tishby, Naftali},
 booktitle = {Advances in Neural Information Processing Systems},
 editor = {Y. Weiss and B. Sch\"{o}lkopf and J. Platt},
 pages = {},
 publisher = {MIT Press},
 title = {Query by Committee Made Real},
 url = {https://proceedings.neurips.cc/paper_files/paper/2005/file/340a39045c40d50dda207bcfdece883a-Paper.pdf},
 volume = {18},
 year = {2005}
}

@inproceedings{
sener2018active,
title={Active Learning for Convolutional Neural Networks: A Core-Set Approach},
author={Ozan Sener and Silvio Savarese},
booktitle={International Conference on Learning Representations},
year={2018},
url={https://openreview.net/forum?id=H1aIuk-RW},
}

@inproceedings{hoi_batch_2006,
	address = {Pittsburgh, Pennsylvania},
	title = {Batch mode active learning and its application to medical image classification},
	isbn = {9781595933836},
	url = {http://portal.acm.org/citation.cfm?doid=1143844.1143897},
	doi = {10.1145/1143844.1143897},
	language = {en},
	urldate = {2024-03-29},
	booktitle = {Proceedings of the 23rd international conference on {Machine} learning  - {ICML} '06},
	publisher = {ACM Press},
	author = {Hoi, Steven C. H. and Jin, Rong and Zhu, Jianke and Lyu, Michael R.},
	year = {2006},
	pages = {417--424},
}

@article{huang_active_2014,
	title = {Active {Learning} by {Querying} {Informative} and {Representative} {Examples}},
	volume = {36},
	issn = {0162-8828, 2160-9292},
	url = {http://ieeexplore.ieee.org/document/6747346/},
	doi = {10.1109/TPAMI.2014.2307881},
	number = {10},
	urldate = {2024-03-29},
	journal = {IEEE Transactions on Pattern Analysis and Machine Intelligence},
	author = {Huang, Sheng-Jun and Jin, Rong and Zhou, Zhi-Hua},
	month = oct,
	year = {2014},
	pages = {1936--1949},
}

@inproceedings{dasgupta_hierarchical_2008,
	address = {Helsinki, Finland},
	title = {Hierarchical sampling for active learning},
	isbn = {9781605582054},
	url = {http://portal.acm.org/citation.cfm?doid=1390156.1390183},
	doi = {10.1145/1390156.1390183},
	language = {en},
	urldate = {2024-03-29},
	booktitle = {Proceedings of the 25th international conference on {Machine} learning - {ICML} '08},
	publisher = {ACM Press},
	author = {Dasgupta, Sanjoy and Hsu, Daniel},
	year = {2008},
	pages = {208--215},
}

@article{Hsu_Lin_2015, title={Active Learning by Learning}, volume={29}, url={https://ojs.aaai.org/index.php/AAAI/article/view/9597}, DOI={10.1609/aaai.v29i1.9597}, number={1}, journal={Proceedings of the AAAI Conference on Artificial Intelligence}, author={Hsu, Wei-Ning and Lin, Hsuan-Tien}, year={2015}, month={Feb.} }

@inproceedings{NIPS2017_8ca8da41,
 author = {Konyushkova, Ksenia and Sznitman, Raphael and Fua, Pascal},
 booktitle = {Advances in Neural Information Processing Systems},
 editor = {I. Guyon and U. Von Luxburg and S. Bengio and H. Wallach and R. Fergus and S. Vishwanathan and R. Garnett},
 pages = {},
 publisher = {Curran Associates, Inc.},
 title = {Learning Active Learning from Data},
 url = {https://proceedings.neurips.cc/paper_files/paper/2017/file/8ca8da41fe1ebc8d3ca31dc14f5fc56c-Paper.pdf},
 volume = {30},
 year = {2017}
}

@INPROCEEDINGS{9093390,
  author={Su, Jong-Chyi and Tsai, Yi-Hsuan and Sohn, Kihyuk and Liu, Buyu and Maji, Subhransu and Chandraker, Manmohan},
  booktitle={2020 IEEE Winter Conference on Applications of Computer Vision (WACV)}, 
  title={Active Adversarial Domain Adaptation}, 
  year={2020},
  volume={},
  number={},
  pages={728-737},
  doi={10.1109/WACV45572.2020.9093390}}

@InProceedings{10.1007/978-3-540-74958-5_14,
author="Donmez, Pinar
and Carbonell, Jaime G.
and Bennett, Paul N.",
editor="Kok, Joost N.
and Koronacki, Jacek
and Mantaras, Raomon Lopez de
and Matwin, Stan
and Mladeni{\v{c}}, Dunja
and Skowron, Andrzej",
title="Dual Strategy Active Learning",
booktitle="Machine Learning: ECML 2007",
year="2007",
publisher="Springer Berlin Heidelberg",
address="Berlin, Heidelberg",
pages="116--127",
isbn="978-3-540-74958-5"
}

@article{baram_online_2004,
	title = {Online choice of active learning algorithms},
	volume = {5},
	journal = {The Journal of Machine Learning Research},
	author = {Baram, Yoram and Yaniv, Ran El and Luz, Kobi},
	month = may,
	year = {2004},
	pages = {255--291},
}

@inproceedings{
Casanova2020Reinforced,
title={Reinforced active learning for image segmentation},
author={Arantxa Casanova and Pedro O. Pinheiro and Negar Rostamzadeh and Christopher J. Pal},
booktitle={International Conference on Learning Representations},
year={2020},
url={https://openreview.net/forum?id=SkgC6TNFvr}
}

@InProceedings{pmlr-v89-cheung19b,
  title = 	 {Learning to Optimize under Non-Stationarity},
  author =       {Cheung, Wang Chi and Simchi-Levi, David and Zhu, Ruihao},
  booktitle = 	 {Proceedings of the Twenty-Second International Conference on Artificial Intelligence and Statistics},
  pages = 	 {1079--1087},
  year = 	 {2019},
  editor = 	 {Chaudhuri, Kamalika and Sugiyama, Masashi},
  volume = 	 {89},
  series = 	 {Proceedings of Machine Learning Research},
  month = 	 {16--18 Apr},
  publisher =    {PMLR},
  url = 	 {https://proceedings.mlr.press/v89/cheung19b.html},
}

@inproceedings{10.5555/3454287.3455365,
 author = {Russac, Yoan and Vernade, Claire and Capp\'{e}, Olivier},
 booktitle = {Advances in Neural Information Processing Systems},
 editor = {H. Wallach and H. Larochelle and A. Beygelzimer and F. d\textquotesingle Alch\'{e}-Buc and E. Fox and R. Garnett},
 pages = {},
 publisher = {Curran Associates, Inc.},
 title = {Weighted Linear Bandits for Non-Stationary Environments},
 url = {https://proceedings.neurips.cc/paper_files/paper/2019/file/263fc48aae39f219b4c71d9d4bb4aed2-Paper.pdf},
 volume = {32},
 year = {2019}
}

@inproceedings{zhang2023algorithm,
  title={Algorithm Selection for Deep Active Learning with Imbalanced Datasets},
  author={Zhang, Jifan and Shao, Shuai and Verma, Saurabh and Nowak, Robert},
  booktitle={Advances in Neural Information Processing Systems},
  year={2023},
  url={https://proceedings.neurips.cc/paper_files/paper/2023/hash/1e77af93008ee6cd248a31723ce357d8-Abstract-Conference.html}
}

@misc{google_google/active-learning_nodate,
	type = {{GitHub}.},
	title = {Google/{Active}-learning},
	year = {2017},
	url = {https://github.com/google/active-learning},
	journal = {Google/Active-learning},
	author = {Google},
}

@inproceedings{zhan_comparative_2021,
	address = {Montreal, Canada},
	title = {A {Comparative} {Survey}: {Benchmarking} for {Pool}-based {Active} {Learning}},
	isbn = {9780999241196},
	shorttitle = {A {Comparative} {Survey}},
	url = {https://www.ijcai.org/proceedings/2021/634},
	doi = {10.24963/ijcai.2021/634},
	language = {en},
	urldate = {2024-04-22},
	booktitle = {Proceedings of the {Thirtieth} {International} {Joint} {Conference} on {Artificial} {Intelligence}},
	publisher = {International Joint Conferences on Artificial Intelligence Organization},
	author = {Zhan, Xueying and Liu, Huan and Li, Qing and Chan, Antoni B.},
	month = aug,
	year = {2021},
	pages = {4679--4686},
}

@inproceedings{li_contextual-bandit_2010,
	address = {Raleigh North Carolina USA},
	title = {A contextual-bandit approach to personalized news article recommendation},
	isbn = {9781605587998},
	url = {https://dl.acm.org/doi/10.1145/1772690.1772758},
	doi = {10.1145/1772690.1772758},
	language = {en},
	urldate = {2024-04-22},
	booktitle = {Proceedings of the 19th international conference on {World} wide web},
	publisher = {ACM},
	author = {Li, Lihong and Chu, Wei and Langford, John and Schapire, Robert E.},
	month = apr,
	year = {2010},
	pages = {661--670},
}

@inproceedings{
citovsky2021batch,
title={Batch Active Learning at Scale},
author={Gui Citovsky and Giulia DeSalvo and Claudio Gentile and Lazaros Karydas and Anand Rajagopalan and Afshin Rostamizadeh and Sanjiv Kumar},
booktitle={Advances in Neural Information Processing Systems},
editor={A. Beygelzimer and Y. Dauphin and P. Liang and J. Wortman Vaughan},
year={2021},
url={https://openreview.net/forum?id=zzdf0CirJM4}
}

@inproceedings{
Ash2020Deep,
title={Deep Batch Active Learning by Diverse, Uncertain Gradient Lower Bounds},
author={Jordan T. Ash and Chicheng Zhang and Akshay Krishnamurthy and John Langford and Alekh Agarwal},
booktitle={International Conference on Learning Representations},
year={2020},
url={https://openreview.net/forum?id=ryghZJBKPS}
}

@inproceedings{
ash2021gone,
title={Gone Fishing: Neural Active Learning with Fisher Embeddings},
author={Jordan T. Ash and Surbhi Goel and Akshay Krishnamurthy and Sham M. Kakade},
booktitle={Advances in Neural Information Processing Systems},
editor={A. Beygelzimer and Y. Dauphin and P. Liang and J. Wortman Vaughan},
year={2021},
url={https://openreview.net/forum?id=DHnThtAyoPj}
}

@inproceedings{NEURIPS2019_95323660,
 author = {Kirsch, Andreas and van Amersfoort, Joost and Gal, Yarin},
 booktitle = {Advances in Neural Information Processing Systems},
 editor = {H. Wallach and H. Larochelle and A. Beygelzimer and F. d\textquotesingle Alch\'{e}-Buc and E. Fox and R. Garnett},
 pages = {},
 publisher = {Curran Associates, Inc.},
 title = {BatchBALD: Efficient and Diverse Batch Acquisition for Deep Bayesian Active Learning},
 url = {https://proceedings.neurips.cc/paper_files/paper/2019/file/95323660ed2124450caaac2c46b5ed90-Paper.pdf},
 volume = {32},
 year = {2019}
}

@article{shwartz-ziv_tabular_2022,
	title = {Tabular data: {Deep} learning is not all you need},
	volume = {81},
	issn = {15662535},
	shorttitle = {Tabular data},
	url = {https://linkinghub.elsevier.com/retrieve/pii/S1566253521002360},
	doi = {10.1016/j.inffus.2021.11.011},
	language = {en},
	urldate = {2025-02-11},
	journal = {Information Fusion},
	author = {Shwartz-Ziv, Ravid and Armon, Amitai},
	month = may,
	year = {2022},
	pages = {84--90},
}
\bibliographystyle{apalike}


\end{document}